\documentclass[nonacm,sigconf]{acmart}

\AtBeginDocument{%
	\providecommand\BibTeX{{%
			\normalfont B\kern-0.5em{\scshape i\kern-0.25em b}\kern-0.8em\TeX}}}

\setcopyright{acmcopyright}
\copyrightyear{2018}
\acmYear{2018}
\acmDOI{10.1145/1122445.1122456}

\acmBooktitle{Woodstock '18: ACM Symposium on Neural Gaze Detection,
	June 03--05, 2018, Woodstock, NY}
\acmPrice{15.00}
\acmISBN{978-1-4503-XXXX-X/18/06}

\begin{document}

\title{A Simple and Interpretable Predictive Model for Healthcare}

\author{Subhadip Maji}
\email{maji.subhadip@optum.com}
\affiliation{%
  \institution{Optum Global Solutions}
  \city{Bangalore}
  \state{India}
  \postcode{560103}
}

\author{Raghav Bali}
\email{raghavbali@optum.com}
\affiliation{%
  \institution{Optum Global Solutions}
  \city{Bangalore}
  \state{India}
  \postcode{560103}
}

\author{Sree Harsha Ankem}
\email{harsha.ankem@optum.com}
\affiliation{%
  \institution{Optum Global Solutions}
  \city{Hyderabad}
  \state{India}
  \postcode{500081}
}

\author{Kishore V Ayyadevara}
\email{vkishore.ayyadevara@optum.com}
\affiliation{%
  \institution{Optum Global Solutions}
  \city{Hyderabad}
  \state{India}
  \postcode{500081}
}
\renewcommand{\shortauthors}{Maji and Bali, et al.}

\begin{abstract}
Deep Learning based models are currently dominating most state-of-the-art solutions for disease prediction. Existing works employ RNNs along with multiple levels of attention mechanisms to provide interpretability. These deep learning models, with trainable parameters running into millions, require huge amounts of compute and data to train and deploy. These requirements are sometimes so huge that they render usage of such models as unfeasible. We address these challenges by developing  a simpler yet interpretable non-deep learning based model for application to EHR data. We model and showcase our work's results on the task of predicting first occurrence of a diagnosis, often overlooked in existing works. We push the capabilities of a tree based model and come up with a strong baseline for more sophisticated models. Its performance shows an improvement over deep learning based solutions (both, with and without the first-occurrence constraint) all the while maintaining interpretability. 
\end{abstract}



\maketitle

\section{Introduction}
Deep Learning has taken the world by a storm and has become the goto choice for developing solutions in areas such as image processing\cite{touvron2019FixRes, DBLP:journals/corr/HeZRS15}, text processing \cite{2019t5}, and even healthcare\cite{Nguyen_2017,retain}. Usage of RNNs (particularly LSTMs \cite{LSTM} and its variants) to model sequential EHR data for disease prediction has been seen in many recent works\cite{Maragatham, Jagannatha} . Recent advancements in deep learning space through the use of attention\cite{Bahdanau}  has helped in adding interpretability as well to these otherwise sophisticated black-box models. RNNs with Attention in healthcare have also been successfully applied in several works\cite{Choi2016MultilayerRL, Choi2015DoctorAP, Zhang2019ATTAINAT, Guo2019AnID}. \citeauthor{retain} \cite{retain} in their paper titled "RETAIN: An Interpretable Predictive Model for Healthcare using Reverse Time Attention Mechanism" used reversed time attention mechanism to achieve good performance while being clinically interpretable for application to the Electronic Health Records data. \citeauthor{dipole} \cite{dipole} rectified some drawbacks of RETAIN, by introducing bidirectional RNNs over normal RNN based approach to capture both the past and future medical experiences of patients. The research and development into this space is possible due to adoption and availability of EHR data. Several works have highlighted the positive impact of such predictive works towards improving quality of healthcare \cite{jha2009use, chaudhry2006systematic,xiao2018opportunities}.

Deep Learning models showcase exemplary performance, at times even outpacing human counter parts. This boost in performance comes at the cost of compute requirements, training time and volume of training data. In many cases, these costs are prohibitory both financially and otherwise. 

We address these limitations/challenges by proposing a simpler yet interpretable tree based predictive model for healthcare. The following were the major motivations behind this work. First and foremost was to test the performance of non-deep learning approaches. We wanted to develop a competitive baseline which can act as a benchmark for highly parameterised current deep learning approaches. Second was to provide a novel way of preparing sequential EHR data for non-deep learning approaches. Our approach had significant impact on overall model performance. Third, models such as RETAIN\cite{retain} provide an intuitive way to interpreting instance level results. We wanted to apply model agnostic approaches to non-deep learning models and provide similar levels of interpretability. The final motivation was to provide an efficient and easy to deploy alternative to Deep Learning models while providing on-par performance and interpretability.

Our model was tested on multiple EHR datasets, each having at least a 24 month historical timeline. We experimented with different datasets and target disease to ensure generalisable performance metrics. We also cater to first-occurrence prediction, i.e. predicting the first ever incidence of a diagnosis in consideration. This constraint adds additional complexity to the prediction task. Our experiments showcase that our simpler approach improves over deep learning based solutions (both, with and without the first-occurrence constraint) all the while maintaining interpretability. 

\begin{figure}
	\centering
	\includegraphics[width=1.0\linewidth]{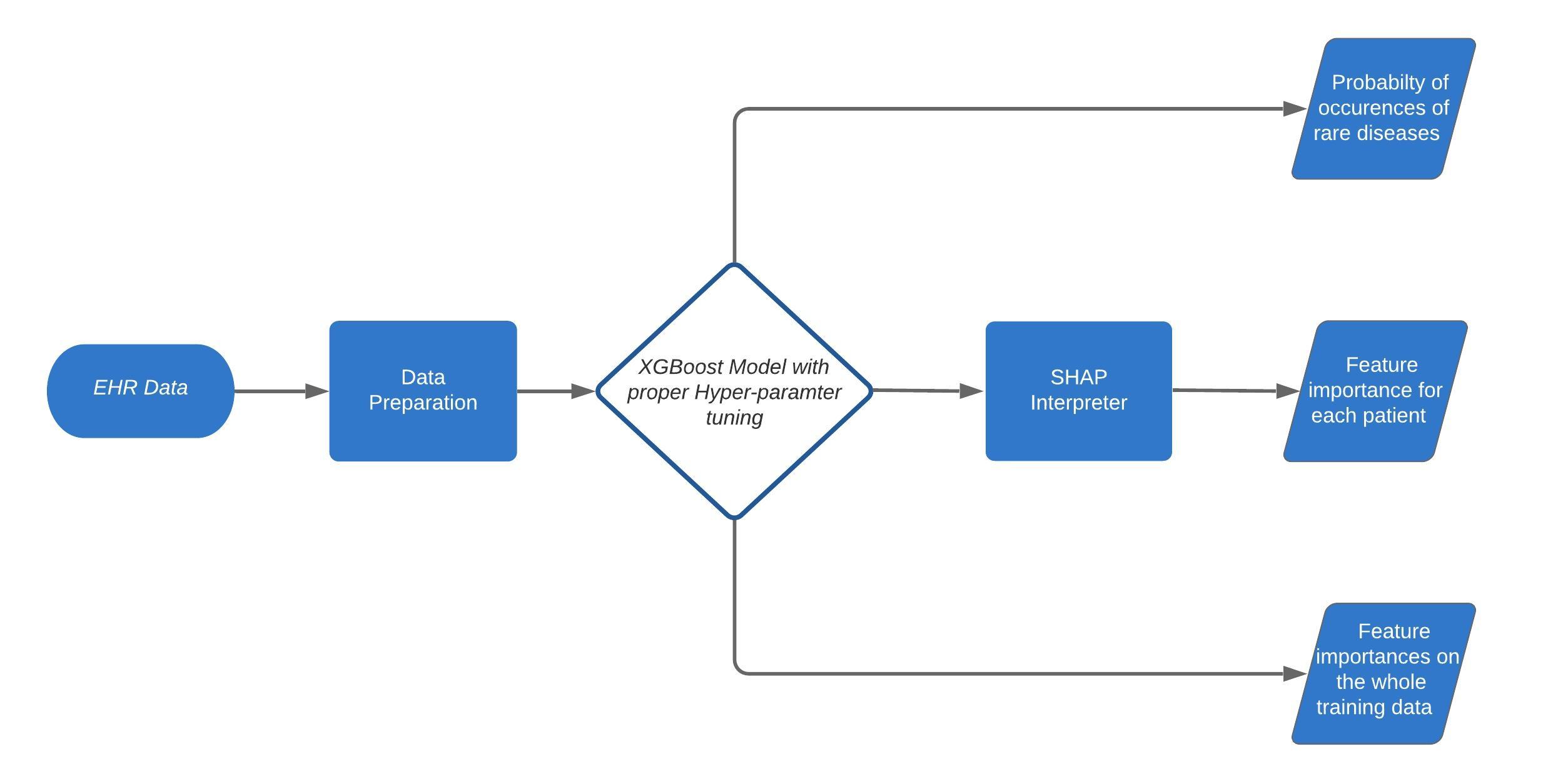}
	\caption{Flow Diagram of the overall approach for rare disease prediction on EHR data using XGBoost and any Model Interpreter}
	\label{fig: architecture}
\end{figure}

The rest of the paper is organised as follows: section 2 details the Data Preparation step. It had a significant impact on the overall model performance. Section 3 describes the overall approach, model choice along with different experiments and their results. We also provide details on the choice of evaluation metric used. We present interesting comparison with different deep learning based approaches. We compare performance with models such as RETAIN\cite{retain}, Dipole\cite{dipole}, etc. In section 4 we discuss the need for model interpretability. We also showcase instance level interpretability results using a model agnostic approach called SHAP\cite{shap}. We also showcase global interpretability results of our model. Section 5 presents commentary on the effectiveness of our work in this domain and section 6 concludes the paper.

\section{Data Preparation}

Data Preparation is an important aspect of this work and a major motivation. As mentioned earlier, the aim was to prepare longitudinal EHR data for first occurrence prediction task. To capture enough historical traits and variability, similar to the works of \citeauthor{retain}, we also pick up a 24 month historical period for training. For the predictions to be useful and actionable, we use a delta of 3 months between the training period and first occurrence date of the diagnosis. Table \ref{tab:data_stat} showcases a quick summary of our EHR dataset.

\begin{table*}
	\centering
	\begin{tabular}{|l|c|c|c|}
		\hline
		& \textbf{Diabetes} & \textbf{Heart Failure} & \textbf{Kidney Failure} \\ \hline
		\# of patients & 261,316 & 84,878 & 85,556 \\ \hline
		\# of medical code groups & 1,840 & 3,327 & 3,327 \\ \hline
		Avg. \# of medical codes per patient & 125.84 & 69.76 & 73.89\\ \hline
		Max \# of medical codes per patient & 424 & 2,278 & 2,278\\ \hline
		Max \# of a medical code per patient & 53 & 617 & 617 \\ \hline
		Max \# of a medical code across all the patients & 845,137 & 2,90,649 & 3,13,662\\ \hline
	\end{tabular}
	\caption{Summary of EHR Datasets utilised for experiments}
	\label{tab:data_stat}
\end{table*}

Let us denote a patient as $p$ having a certain history of diagnosis, denoted as $H=\{t_1,t_2,....t_N\}$ where $t_i$ is one timestep or a visit in his/her history. Each timestep consists of various diagnosis codes (represented as ICD codes), procedures codes (represented as CPT codes), prescription (represented as RX codes) and demographic details of the patient and can be represented as  $t_i=\{ICD_{(1..N)}, CPT_{(1..N)},demographics\}$ . Assume that the task is to predict the first occurrence of a disease $D$. Given that this patient was diagnosed with $d$ at time steps $t_n$ and $t_{(n+p)}$  (where p > 0) time step. The first occurrence of $d$ for $p$ would time step $t_n$ which would be our target instance. The response variable $is\_d$ (for instance $is\_diabetic$) would be set to 1 for such patients and 0 otherwise (i.e. patients who have never been diagnosed with diabetes).

Using the above mentioned procedure, we prepare the response variables for our population. We prepare a different dataset for each diagnosis. The feature space consists of $ICD codes$ along with demographic attributes like $age$ and $gender$. To prepare an aggregated feature vector for each patient $p$, we dissolve the time steps, i.e. the patient vector is represented as:

 \begin{equation}
 p = \{ICD_1..ICD_N, CPT_1,..CPT_N,RX_1..RX_N,age,gender\}
\end{equation}
 
 , where the value for each $ICD_i$ depends upon the experiment being considered. We experimented with two different versions. First experiment involved setting the value of each $ICD_i$, $CPT_i$ and $RX_i$ to the number of such diagnosis, procedures or medication. So, for a given patient $p_i$  the feature vector would be referred as:

\begin{equation}
 p_i = \{X_i, y_i\} : \{(n_{11}, n_{12}, ....., a_1, g_1, ...), y_i\}
\end{equation} 
 
 , where $n_{ij}$ is the ICD count value for patient \texttt{i} and ICD \texttt{j} (similarly for CPT and RX codes), $a_i, g_i$ and $y_i$ are the age, gender and response value respectively for patient \texttt{I}; with $a_i \in (0, \infty)$ $g_i \in$ \{Male, Female\} and $y_i \in \{0, 1\}$. In the second experiment we treated $ICD_i$ as a binary categorical feature. We share details on the model performance on these two different experiments in the following section.

\section{Experimental Setup}

Deep learning models are very effective in a majority of tasks. Before the widespread usage of such models, tree based models \cite{doi:10.1002/cyto.990080516, xgboost,randomForest} were the go-to choice. The major reasons behind the popularity of tree based models were their low bias, robustness against outliers, ease of interpretability and speed of training and inference. These were our motivations as well to decide upon XGBoost\cite{xgboost}  as our model of choice. Being battle-tested in different scenarios such as production use-cases, academics and ML competitions further reinforced our decision. Global interpretability is another factor which contributes in understanding model behaviour. 
We utilised different model-agnostic instance level feature interpreters like LIME\cite{lime} and SHAP\cite{shap} as well. These instance level approaches were used to identify contributing features at each patient level. Figure \ref{fig: architecture} shows our overall approach in a flow diagram.  

\subsection{Evaluation Metric}
Disease incidence is usually very low in observed patient samples. This low incidence leads to issues associated with class imbalance for classification models. Accuracy as a measure is not helpful in such scenarios and leads to models biased towards majority class. We use Receiver Operating Characteristic-Area Under Curve or ROC-AUC\cite{roc_auc} as our metric of choice. ROC is a probability curve to understand performance measurement of the model at different thresholds while the area under the curve denotes the degree of separability between classes. ROC-AUC is robust measure for imbalanced datasets. We also measure \textit{Recall@K} as an additional metric and define it as: given the predicted probability scores across all the observations binned into deciles, \textit{Recall@30} is defined as the percentage of true cases for which the predicted probability falls in the top 3 deciles.

\subsection{Experiments and Results}
We performed various experiments to understand model performance using ROC-AUC as our metric of choice. The aim is to prepare a classifier to identify first occurrence of a disease in the dataset. Considering our focus is towards a classification task (first occurrence/diagnosis for diabetes, heart failure and kidney failure respectively), we split the dataset into three parts: $train$, $validation$ and $test$. Our datasets are highly imbalanced. The class imbalance stands at (\texttt{class 1:class 0 = $12$:$88$ }) for diabetes, (\texttt{class 1:class 0 = $15$:$85$ }) for  heart failure and (\texttt{class 1:class 0 = $14$:$86$ }) for kidney failure . Stratified sampling was performed while splitting the dataset into $train$, $validation$ and $test$ to maintain class distribution.

As a first step we fit a logistic regression model on our datasets. This was done to have a baseline in place and understand the relative strength of each of the models. Since this was a binary classification task, we could directly fit a logistic regression model. The models achieved an ROC-AUC of $0.711$, $0.754$ and $0.731$ for diabetes, heart failure and kidney failure respectively. This is quite a decent performance given the simplicity of the model. The models where we utilised count of features rather than binary categorisation achieved better performance throughout. For the rest of this section we will refer to count based feature set as our primary dataset unless stated otherwise.

Moving ahead with this baseline, the next experiment involved fitting an XGBoost model with default parameters. XGBoost is a tree based boosting algorithm with numerous hyper-parameters such as learning rate, number of estimators, regularisation parameters and so on. We denote this XGBoost model with default parameters as $xgb_{def}$ henceforth. The XGBoost models with default setting for diabetes, heart failure and kidney failure resulted in an ROC-AUC value of $0.78$, $0.837$ and $0.823$ respectively, which is good improvement over the logistic regression baseline.

\begin{table*}[]
\begin{tabular}{lc|c|c|c|l|}
\cline{3-6}
\textbf{}                                & \multicolumn{1}{l|}{\textbf{}} & \multicolumn{4}{c|}{\textbf{Optimal Values}} \\ \hline
\multicolumn{1}{|l|}{\textbf{Important Parameters}} &
  \textbf{\textbf{Default Values}} &
  \textbf{\textbf{Diabetes}} &
  \textbf{\textbf{Heart Failure}} &
  \textbf{\textbf{Kidney Failure}} &
  \textbf{Heart Failure-ICD-Full} \\ \hline
\multicolumn{1}{|l|}{learning\_rate}     & 0.1                            & 0.01      & 0.01      & 0.01      & 0.01     \\ \hline
\multicolumn{1}{|l|}{n\_estimators} &
  100 &
  6674 &
  {\color[HTML]{000000} 7059} &
  {\color[HTML]{000000} 4795} &
  5733 \\ \hline
\multicolumn{1}{|l|}{max\_depth}         & 3                              & 4         & 4         & 5         & 4        \\ \hline
\multicolumn{1}{|l|}{min\_child\_weight} & 1                              & 6         & 1         & 4         & 1        \\ \hline
\multicolumn{1}{|l|}{gamma}              & 0                              & 0.4       & 0         & 0         & 0.1      \\ \hline
\multicolumn{1}{|l|}{reg\_alpha}         & 0                              & 1e-5      & 6         & 1         & 1        \\ \hline
\multicolumn{1}{|l|}{reg\_lambda}        & 1                              & 100       & 1         & 10        & 1        \\ \hline
\multicolumn{1}{|l|}{subsample}          & 1                              & 0.9       & 0.95      & 0.8       & 0.7      \\ \hline
\multicolumn{1}{|l|}{colsample\_bytree}  & 1                              & 0.9       & 0.45      & 0.6       & 0.6      \\ \hline
\end{tabular}
\caption{Comparison between default and optimal hyper-parameter values of XGboost for each target disease}
\label{tab:default_and_best_hyperparameter}
\end{table*}
\subsection{Hyper-parameter Tuning}
As mentioned earlier, XGBoost has a host of hyper-parameters available for fine-tuning. Since our aim was to try and push the boundaries of non-deep learning models, it was logical next step to fine tune the $xgb_{def}$ model. One of the ways of identifying the right values for each of the hyper-parameters is to perform a greedy search. We did not proceed with the usual grid-search due to the shear size of the hyper-parameter search space. A grid search would have been too time and effort consuming.

The greedy search paradigm works as follows:
\begin{itemize}
	\item Use $xgb_{def}$ model as the base for this greedy search
	\item Let the rank ordered list of hyper-parameters be denoted as \break
	$H=\{learning\_rate,n\_estimators,....,reg\_lambda\}$.
	\item Rank order hyper-parameters based on their importance. 
	\item For each $h_i$ $\in$ $H$ :
	\begin{itemize}
		\item Fit XGBoost on $validation$ dataset to identify optimal value of $h_i$ keeping all other hyper-parameters in $H$ as constant.
		\item Update optimal value of $h_i$ in $H$
	\end{itemize}
\end{itemize}

\begin{figure*}[]
	\centering
	\includegraphics[width=0.9\linewidth]{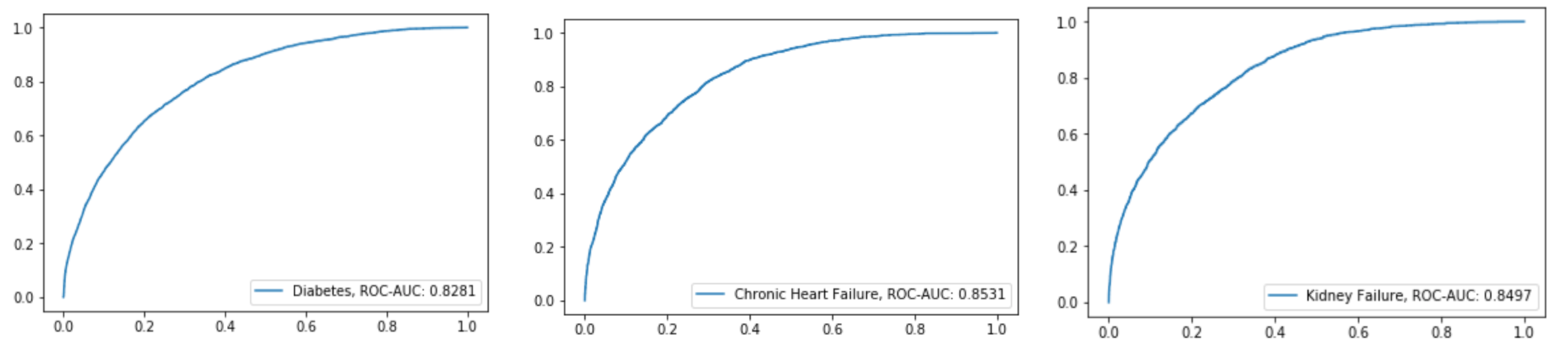}
	\caption{ROC-AUC plot for fine tuned XGBoost Model ($xgb_{opt}$) on EHR Data}
	\label{fig:auc_roc}
\end{figure*}

The above process helped us in achieve optimal values for each of our hyper-parameters in consideration. The optimal parameter values are mentioned in table \ref{tab:default_and_best_hyperparameter} for reference\footnote{only those hyper-parameters are listed in the table which changed their values from default settings. The nomenclatures of the hyper-parameters are from \texttt{xgboost.XGBClassifier\cite{xgboost_python}}}. The fine-tuned XGBoost (denoted as $xgb _{opt}^{diabetes}$) shows an improvement of approximately $6.5\%$ over $xgb_{def}$ with an ROC-AUC of $0.8281$ for diabetes. The ROC-AUC plot is shown in the Figure \ref{fig:auc_roc}. We observed similar lifts for heart failure and kidney failure models as well. The ROC-AUC for $xgb_{opt}^{heart}$  and $xgb_{opt}^{kidney}$ stand at $0.853$ and $0.849$ respectively.

\begin{table*}[]
\centering
\begin{tabular}{|l|c|l|l|l|l|l|}
\hline
\textbf{}      & \multicolumn{2}{c|}{\textbf{\textbf{Diabetes}}} & \multicolumn{2}{c|}{\textbf{Heart Failure}} & \multicolumn{2}{c|}{\textbf{Kidney Failure}} \\ \hline
\textbf{Model} & \textbf{ROC-AUC}     & \textbf{Recall@30}    & \textbf{ROC-AUC}   & \textbf{Recall@30}  & \textbf{ROC-AUC}   & \textbf{Recall@30}   \\ \hline
RETAIN              & $0.7831 \pm 0.0062$                      & 0.2998 & $0.8385 \pm 0.002$  & 0.4047 & $0.8202 \pm 0.001$  & 0.3312 \\ \hline
Dipole              & $0.7901 \pm 0.0085$                      & 0.3002 & $0.8291 \pm 0.009$  & 0.3998 & $0.8211 \pm 0.008$  & 0.3357 \\ \hline
Logistic Regression & $0.7113 \pm 0.0000$                      & 0.2885 & $0.7543 \pm 0.0000$ & 0.3765 & $0.7319 \pm 0.0000$ & 0.3129 \\ \hline
$xgb_{base}$        & \multicolumn{1}{l|}{$0.78 \pm 0.000$}    & 0.3047 & $0.8377 \pm 0.0002$ & 0.3975 & $0.8233 \pm 0.0001$ & 0.3301 \\ \hline
$xgb_{oneHot}$      & \multicolumn{1}{l|}{$0.805 \pm 0.0003$}  & 0.3052 & $0.8404 \pm 0.0000$ & 0.4002 & $0.8258 \pm 0.0004$ & 0.3311 \\ \hline \hline
$xgb_{opt}$         & \multicolumn{1}{l|}{$0.8281 \pm 0.0005$} & 0.3093 & $0.8531 \pm 0.0005$ & 0.4113 & $0.8497 \pm 0.0007$ & 0.3374 \\ \hline
\end{tabular}
\caption{Comparison of results of our proposed method with recent papers}
\label{tab:result_compare}
\end{table*}


\subsection{Comparison with Recent works}
Works by \citeauthor{retain}\cite{retain} and \citeauthor{dipole}\cite{dipole} utilise complex attention mechanisms to showcase improvements in their results. These works compare their results against weak baselines only. These works also overlook experiments concerning first occurrence prediction. We believe this additional constraint is an important one for the models to be useful in real life use cases. We also observed that the performance (across models) tends to improve drastically if this constraint is removed. This makes intuitive sense as for many diagnosis, a repeat occurrence is quite obvious. From a business and healthcare stand-point, it makes sense to predict first occurrence to take any preventive/corrective action in time.

To provide a common framework, competitive baseline and useful constraints, we trained RETAIN\cite{retain}\footnote{We used the code from  \href{https://github.com/mp2893/retain}{https://github.com/mp2893/retain}} and Dipole\cite{dipole}\footnote{Among the three attention layers described, General Attention layer worked best} on our datasets, preparing data in the formats expected and performed hyper-parameter tuning to report the best results on our test dataset. 

We observed significant improvements in ROC-AUC values for both RETAIN and Dipole as compared to $xgb_{def}$ and logistic regression baselines. This was expected as both these models are highly parameterised and complex implementations. Also, both these works present improvements against logistic regression in their respective works as well. The surprising aspect was the comparison with our fine-tuned XGBoost model, i.e. $xgb_{opt}$. Our proposed model shows considerable improvements as compared to RETAIN and Dipole for all three target diseases respectively. The results are showcased in table \ref{tab:result_compare}  for reference.

\subsection{Experiments with Full ICD Feature Set}

The experiments and results outlined in previous section utilised ICD codes truncated till 3 characters (or ICD3 for short). For instance, diagnosis code $250.31$ refers to \textit{Diabetes with other coma, type I [juvenile type], not stated as uncontrolled}. We truncate the same to only $250$ which refers to the class of diabetic diagnosis. By doing so, we reduce the overall dimensionality of our already sparse feature set. 

Even though such a grouping is helpful in reducing the impact of a sparse feature set, it leads to loss of understanding/interpretability. To enable better and granular interpretation, we experimented with complete ICD codes or ICD-Full \footnote{For our experiments, ICD-Full refers to complete codes and ICD3 refers to only 3 digit codes. Do not confuse this with version number of ICD codes.}. The dataset was prepared as mentioned in the Data Preparation section with only difference being the feature set consists of ICD-Full while targets are still ICD3. This was done to ensure we have enough training samples for each class. This new dataset was used to train and tune RETAIN, Dipole and Xgboost for comparison. Similar to previous experiments, in this case also XGBoost outperformed its more sophisticated competitors on ROC-AUC metric. Results are shared in table \ref{tab:result_icd5} for reference. We attribute the improvement in performance across models to the added granularity in the feature set all the while maintaining similar class distribution. The results were cross validated to ensure model stability.

\begin{table*}[]
\centering
\begin{tabular}{|l|l|}
\hline
\multicolumn{1}{|c|}{\textbf{Models}} & \multicolumn{1}{c|}{\textbf{ROC-AUC}} \\ \hline
RETAIN      & $0.8539 \pm 0.001$ \\ \hline
Dipole      & $0.8561 \pm 0.02$  \\ \hline
$xgb_{opt}$ & $0.8626 \pm 0.007$ \\ \hline
\end{tabular}
\caption{Heart Failure Prediction using ICD5 in the feature set.}
\label{tab:result_icd5}
\end{table*}

\section{Interpretability}

Interpretability is an important factor when it comes to use cases such as disease prediction. Typically there is a trade-off between model performance and its interpretability. Most Deep Learning models are highly complex and are often treated as black boxes. To overcome these limitations, works by \citeauthor{retain}\cite{retain} and \citeauthor{dipole}\cite{dipole} utilise attention mechanisms.

\begin{figure*}[]
	\centering
	\includegraphics[width=0.9\linewidth]{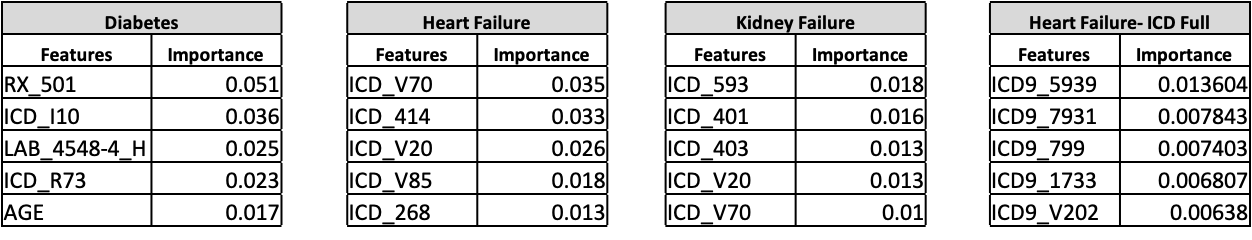}
	\caption{Interpreting Patient Level Predictions. We used SHAP to understand features impacting model prediction probabilities. Here we have 3 randomly chosen patients for diabetes, heart failure and kidney failure prediction tasks.}
	\label{fig: global_feat_imp}
\end{figure*}

Since our fine-tuned XGBoost model, $xgb_{opt}$ is not a deep learning model, the interpretability at instance level had to be solved in a different way. For global (or dataset) level feature importance, tree based algorithms are a go-to choice. The \texttt{XGBClassifier}\cite{xgboost_python} also provides similar functionality out of the box. Important features for $xgb_{opt}$ for first occurrence prediction of diabetes are reported as ICD\_I10(Hypertension), ICD\_R73(Elevated blood glucose levels), etc. which are inline with factors leading to diabetes. Heart Failure task using ICD-Full as feature set resulted in top 5 features as ICD$5939$ (Unspecified disorder of kidney and ureter), ICD$7931$ (Nonspecific (abnormal) findings on radiological and other examination of lung field).Figure \ref{fig: global_feat_imp} presents the top 5 features in detail for each of the target diseases.

\subsection{Patient Diagnosis Interpretability}
$xgb_{opt}$ outperforms its deep learning counterparts for the task of first occurrence prediction while remaining globally interpretable. One downside of XGBoost is its inability to provide instance or in this case, patient level interpretability. To handle this scenario, we leverage a model agnostic approach by \citeauthor{shap}\cite{shap}. 

\begin{figure*}[]
	\centering
	\includegraphics[width=0.9\linewidth]{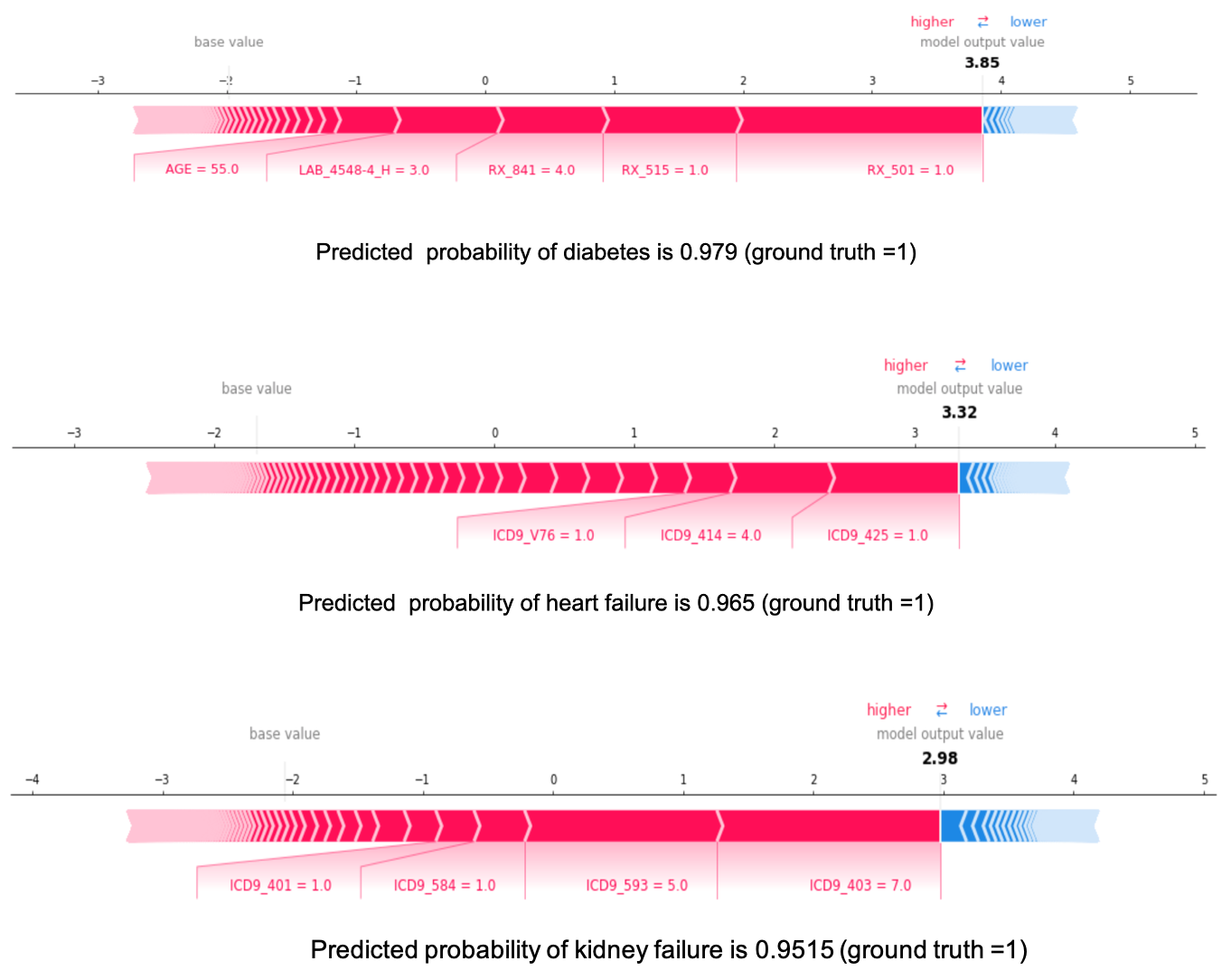}
	\caption{Interpreting Patient Level Predictions. We used SHAP to understand features impacting model prediction probabilities. Here we have 3 randomly chosen patients for diabetes, heart failure and kidney failure prediction tasks.}
	\label{fig: shap}
\end{figure*}

This approach mimics the behaviour outlined in the works of \citeauthor{retain}. Their work explains theoretical motivations and working in detail. To better understand the impact on our work, let us work through an instance of a patient from our test dataset itself. Let us consider a randomly sampled patient from our test dataset for diabetes. We use $xgb_{opt}$ to predict the first occurrence probability of this patient being diabetic. This particular patient turns out to be diabetic with a probability score of 0.979 (ground truth for this patient was observed to be 1). XGBoost is supported by the SHAP framework out of the box. Upon analysing this particular instance using SHAP, we observe the following for this particular patient. Features such as age, LAB\_4548-4\_H (a diagnostic test for Haemoglobin A1c), RX\_841(diabetes testing supplies) and so on have positive SHAP values. Positive SHAP values move the logit value (classification decision) of the classifier from approximately $-3.0$ to $3.85$. These are the past diagnosis, events, lab test or prescriptions which the model uses to have a high probability ($0.979$) for predicting this patient as diabetic three months down the line. The same is visually showcased for heart failure and kidney failure as well in figure \ref{fig: shap} for reference.

Similar to attention based interpretation plots as showcased in RETAIN\cite{retain}, we leverage SHAP values and force plots to provide patient level interpretability for our work. Similar exercise can be performed using LIME\cite{lime} to understand instance level feature importance.

\subsection{Known Limitations}
One known limitation of XGBoost models as compared to deep learning counterparts\cite{retain}\cite{dipole} is the visit-level importance. In the data preparation step, we outlined the fact that we the aggregated feature vector does not include time aspect of a patient's history. We dissolve the time steps while preparing the patient vector $p$.

While sequence to sequence based deep learning models can provide time-step level interpretability, our model does not have such capability out of the box. To handle this scenario, we present a simple workaround. We firstly narrow down the top most important features at the patient level. The next step is to identify visits which had these features present. We can then mark such visits as important in identifying the final diagnosis and also provide physicians with supplementary information regarding such a decision.

\section{Effectiveness of Tree Based Approaches}
Our experiments outline the effective predictive performance of XGBoost based models as compared to its sophisticated deep learning counterparts. Despite having far less parameters, our optimised versions were able to outperform attention based architectures such as RETAIN\cite{retain} and Dipole\cite{dipole}.

Such a strong baseline can be attributed to two main aspects of our experimental setup. The first and the foremost is the domain and its data. We shared our results and corresponding interpretations with medical professionals. The experts were able to verify the results and the interpretations from a random sample of our test sets. They also highlighted the importance of sequential/longitudinal nature of electronic health records. Though an important factor in a number of diagnosis (such as Alzheimer's ), not all diagnosis are time dependent, especially for the target diseases in our experiments. They highlighted the fact that even though past diagnosis impact current and future health states, the time gap is not always an important factor. This goes hand in hand with our results and the fact that a simpler model out performs complex ones. This also highlights a gap in the data recording process. Each diagnosis in EHR datasets is associated with a visit to a doctor/medical professional and is not the actual date of incidence. Thus, the time information from EHR dataset is dependent upon when a particular person visits a medical facility for diagnosis. This might include delays due to personal preferences such as seriousness of symptoms, access to healthcare, pain tolerance and so on. Such variability between incidence and reporting requires more study and experiments. 

The second aspect is from the algorithmic standpoint. Despite successful application across various domains and data types (mostly unstructured), deep learning is yet to make a mark when it comes to tabular or structured datasets. Tree based ensembles, especially XGBoost and variants dominate this space\cite{Kaggle}. Real world datasets are typically high dimensional yet sparse in nature. In other words, they can be represented in a lower dimensional space easily (say a hyperplane). This process is termed as unfolding or manifold learning. Tree based boosting algorithms are highly efficient for manifold learning with hyperplane boundaries (a characteristic of tabular datasets) \cite{criminisi2012decision}. Another reason behind better performance of tree based ensembles over deep learning counterparts is their ease and speed of training. Deep Learning models are over parameterised and even though they are termed as universal function approximators, finding the optimal set of parameters is not a trivial task. These require far more training samples and time as compared to traditional methods\cite{wolpert1997no}.

\section{Conclusion}
We presented a simple and interpretable predictive model for disease prediction. Sophisticated and complex deep learning models are the focus of research work in disease prediction domain.\citeauthor{retain}\cite{retain} present attention based approach to prepare interpretable disease prediction model. Their work and the likes present comparison with weak baselines, mostly using logistic regression. The focus of this work was to push the capabilities of a tree based non-deep learning model and come up with a strong baseline for more sophisticated models. We present a novel data preparation pipeline which is observed to have a positive impact on the overall model performance. We used ROC-AUC\cite{roc_auc} as our evaluation metric, given the fact that dataset in consideration is highly skewed. Our work outlined different experiments and a simple algorithm to fine-tune the XGBoost model for performance. We compared the performance of our work with that of RETAIN\cite{retain} and Dipole\cite{dipole}. It was surprising to observe that our fine-tuned model outperformed these deep learning solutions by a good margin. This was despite the fact that both deep learning implementations were fine-tuned with respect to the dataset in consideration. We also presented strategies to interpret our model at both global and instance levels. The instance level interpretation utilised SHAP framework by \citeauthor{shap}. SHAP values help us understand patient level feature importance. We also discussed about the limitation of our model while identifying visit level importance. We closed by providing a simple workaround for this known limitation. We leveraged XGBoost implementation by \citeauthor{xgboost}\cite{xgboost} to prepare our models.

\section*{Acknowledgements}
We would like to thank Vineet Shukla and Saikumar Chintareddy for helpful discussions and inputs to improve the solution, and the whole diagnosis prediction team for their contributions.

\bibliographystyle{ACM-Reference-Format}
\bibliography{sample-sigconf}
\end{document}